# Training Dynamic based data filtering may not work for NLP datasets

**Arka Talukdar**[*]
at4786@nyu.edu
New York University

**Monika Dagar**[*]
md4676@nyu.edu
New York University

**Prachi Gupta**[*]
pg1647@nyu.edu
New York University

**Varun Menon**[*]
vk21486@nyu.edu
New York University

## Abstract

The recent increase in dataset size has brought about significant advances in natural language understanding. These large datasets are usually collected through automation (search engines or web crawlers) or crowdsourcing which inherently introduces incorrectly labeled data. Training on these datasets leads to memorization and poor generalization. Thus, it is pertinent to develop techniques that help in the identification and isolation of mislabelled data. In this paper, we study the applicability of the Area Under the Margin (AUM) metric to identify and remove/rectify mislabelled examples in NLP datasets. We find that mislabelled samples can be filtered using the AUM metric in NLP datasets but it also removes a significant number of correctly labeled points and leads to the loss of a large amount of relevant language information. We show that models rely on the distributional information instead of relying on syntactic and semantic representations.

## 1 Introduction

Modern deep learning networks are becoming deeper and powerful, and have led to significant advances in Natural Language Processing (NLP) (Devlin et al., 2019), Computer Vision (He et al., 2015), and Speech Processing (Graves et al., 2013). However, these networks rely on large labeled datasets to be effective.

The creation of large labeled datasets has fueled the advances in NLP (Rajpurkar et al., 2016; Bowman et al., 2015). Abundant labeled data increases the likelihood of learning diverse phenomena, which in turn leads to models that generalize well (Linzen, 2020).

Curating expert annotated datasets is very time-consuming and costly (Malik and Bhardwaj, 2011) therefore large language datasets are usually collected through crowd-sourcing, by hiring human annotators (Wang et al., 2019) or by crawling the web. Such methods inherently introduce label noise in the resulting data. Mislabelled training data is particularly problematic for deep neural networks with billions of parameters because they can overfit on the mislabelled data and achieve zero training error even on randomly assigned labels (Zhang et al., 2016). Training models with noisy labels can also lead to misclassification on easy examples during test-time (Beigman and Beigman Klebanov, 2009).

It is prohibitively costly to manually remove mislabeled samples from large datasets. Hence, the need arises to create an automated pipeline to analyze and clean datasets. Area Under the Margin (AUM) metric was designed to identify and eliminate noisy data. It can be used as a plug-and-play method within the training pipeline of any classification network with minimal overhead (Pleiss et al., 2020). AUM shows promising results in identifying mislabelled samples in image classification datasets.

Thus, in this paper, we investigate the applicability of the AUM metric on text classification datasets. We make the following contributions: (i) We show that the AUM metric has lower efficacy for filtering mislabelled data in NLP datasets than image datasets. (ii) We hypothesize that AUM does not work as expected in NLP datasets as it did in image datasets because of the intrinsic nature of the data samples. They have high intra-class and low inter-class feature similarity (Ho et al., 2021), which is usually not the case in NLP datasets. We show samples from NLP datasets to corroborate our hypothesis.

---

[*] Equal contribution

## 2 Related Work

**Detecting mislabelled Instances.** Pleiss et al. (2020) and Swayamdipta et al. (2020), both use model behavior on each sample over the training process, also known as training dynamics, to identify mislabelled instances in classification datasets but using different metrics. For each sample, Pleiss et al. (2020) finds the difference in logit value of the assigned class (gold label) and the highest other logit value among the non-assigned classes averaged over training epochs called the area under the margin (AUM) metric. They also introduce a fake class with samples having only mislabelled instances by definition to find a threshold AUM value. Samples with low AUM scores are likely to be mislabelled and the threshold AUM value is used to filter out such mislabelled instances. Swayamdipta et al. (2020) uses mean and standard deviation of the gold label probabilities over the training epochs, called confidence and variability scores, respectively, for each sample. They classify samples with low confidence and low variability as either mislabelled or "hard-to-learn" for the model.

Bhardwaj et al. (2010) uses statistical methods to find annotators whose annotations differ considerably from the remaining annotators and use manual inspection to decide the verdict for samples annotated by these annotators. Müller and Markert (2019) classifies training samples with the lowest gold label probabilities on a robust classifier as potentially mislabelled followed by their manual review for final decision. Zhang and Sugiyama (2021) detects samples with erroneous labels using an instance-dependent noise model along with instance-based embedding to capture instance-specific label corruption.

**Learning in the Presence of Noisy Labels.** Several efforts have been made to account for noise in the data and prevent the model from memorizing wrong examples without actually identifying and removing such examples from the training set. Li et al. (2020) replaces the last layer of models trained on noisy data with a linear layer trained on a small set of clean data, Jindal et al. (2019) adds a non-linear noise-modeling layer on top of the target text-classification model. Kang and Hashimoto (2020) improves faithfulness in text generation by adaptively removing high log loss examples during the training process.

| Percentile | Acc. on Unfiltered Data | Acc. on Filtered Data |
|---|---|---|
| 1 | 87.91 ± 0.25 | 88.02 ± 0.79 |
| 10 | 87.79 ± 0.50 | 87.79 ± 0.61 |
| 50 | 87.87 ± 0.36 | 87.72 ± 0.48 |
| 90 | 87.79 ± 0.49 | 87.72 ± 0.48 |

Table 1: Accuracy on SST-2 dataset at different AUM threshold percentiles; Sieving

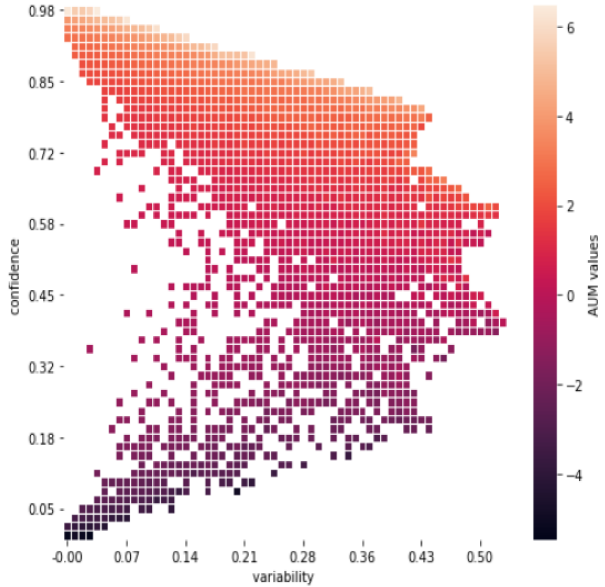

Figure 1: Visualizing AUM values of SST-2 samples along with their Data Map plotted as per Swayamdipta et al. (2020) shows considerable overlap in samples with low AUM values and samples identified as hard-to-learn/mislabelled as per Data map (sampled with low confidence and low variability scores).

**Other types of Noise in text data.** Depending on the type of supervisory signal and data acquisition method, language datasets can have noise in the form other than labeling errors like spelling errors, grammatical errors (Subramaniam et al., 2009; Malik and Bhardwaj, 2011). Caswell et al. (2021) provides large-scale systematic quality analysis of various web-crawled multilingual datasets and found large amounts of samples with inconsistencies in language codes and mistranslations. Robust-to-noise word embeddings (Malykh, 2019), noisy data classifiers trained on clean data, and synthetically generated noisy data (Xu and Koehn, 2017) are some efforts to deal with non-label noise in language data. In this work, however, we only study label noise.

**Relation to annotator disagreement.** The work on dealing with noisy labels in classification datasets can also be related to the work on studying annotator disagreements. Previous work (Beigman and Beigman Klebanov, 2009; Beigman Klebanov et al., 2008; Pavlick and Kwiatkowski, 2019) shows that there can be two reasons for disagreements in annotator labels in crowdsourced datasets: difference of opinion and attention slip. Former generally occurs when different groups of people agree with a different assigned label for a sample based on their understanding of the text. Latter generally occurs due to attention slip or genuine mistake during annotation. As a direction of future work, comparing samples identified as mislabelled using the AUM method with samples that get relatively low agreement among crowd worker annotators can provide meaningful insights.

| Sample Id | Text | Label |
|---|---|---|
| 390 | He I often sees Mary. | 1 |
| 5766 | Heidi believes any description of herself. | 1 |
| 2801 | Paula hit the sticks. | 0 |
| 1522 | That the sun is out was obvious. | 0 |
| 8332 | I wanted Jimmy for to come with me. | 1 |

Table 2: Filtered examples from CoLA dataset (1 = grammatically acceptable; 0 = grammatically unacceptable)

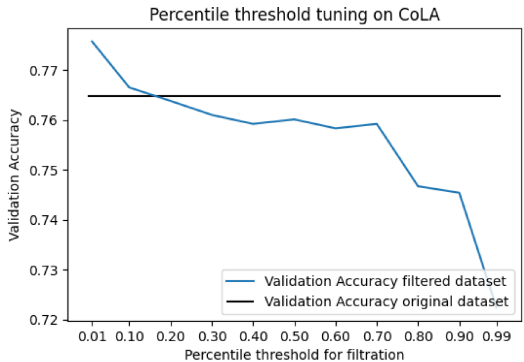

Figure 2: CoLA: Percentile threshold vs Validation accuracy

## 3 Implementation Details

### 3.1 Filtering data with AUM

We use the AUM metric and methodology from Pleiss et al. (2020) to identify training samples with AUM values below a threshold value as mislabelled. To calculate this threshold, original training data is distributed to make a fake class with equal samples from all the original classes. The classification model is trained on this new dataset configuration to generate the AUM values for all the data points. Samples in fake class are by definition mislabelled hence AUM values of fake class samples can be used as a threshold for the samples in the original classes. This method is repeated to find the mislabelled samples among the samples which were in fake class initially. In the second run, a fake class is created such that it does not have any samples which were in the fake class in the first run.

As discussed in Section 5, we observed that the heuristic-based thresholding technique suggested in Pleiss et al. (2020), wherein they used the AUM value of the 99th percentile threshold sample as the threshold to filter the data, does not show major improvement in NLP datasets. We thus consider the AUM threshold value as a hyperparameter and fine-tune it. We also propose a method to rectify the labels and reuse the data for training (discussed in Section 4).

### 3.2 Experimental Setup

We finetuned a distillBERT-base model on SST-2 (Socher et al., 2013) and CoLA (Warstadt et al., 2019), pretrained using a masked language modeling (MLM) objective (Sanh et al., 2019) with a default AdamW optimizer (Loshchilov and Hutter, 2017). We selected distillBERT for our experimentation because it is small and fast while preserving over 95% of BERT's performance measured on GLUE benchmark (Sanh et al., 2019).

## 4 Experiments

Following the recommendations from Pleiss et al. (2020), we test the efficacy of AUM on synthetic-noisy and real-world NLP datasets to identify mislabelled samples. To create synthetic-noisy datasets, we injected noise in the real-world datasets by uniformly sampling data points and flipping their labels. We run two experiments on both types of datasets. First, we discard the samples classified by AUM as mislabelled; we will refer to this process as Sieving. Second, instead of discarding samples, we rectify the label and reuse them for training; we will refer to this process as Flipping. Since we train on binary-classification tasks, we flip the label of the samples which are classified as mislabelled.

As noted in Pleiss et al. (2020), the AUM threshold for filtration is dataset dependent. The authors provide a simple heuristic for classifying samples as mislabelled; samples with AUM lower than the 99th percentile threshold sample's AUM will be classified as mislabelled. Further, they also note that the filtration performance is robust to this hyperparameter (percentile threshold). In our exper-iments with CoLA and SST-2, we observed that this does not translate well to NLP datasets. Indiscriminately removing samples with AUM less than the 99th percentile threshold consistently had poor performance compared to the unfiltered dataset. As the percentile threshold was reduced, the validation accuracy increases as seen in Figure 2 for CoLA and Table 1 for SST-2. It is because that samples with AUM lower than a high percentile threshold could be a hard-to-learn sample which helps generalization (Swayamdipta et al., 2020). To support this hypothesis, we classified data points using the method (refer to Section 2 for details) suggested by Swayamdipta et al. (2020). We observed significant overlap in samples identified as mislabelled using AUM and samples identified as hard-to-learn using Swayamdipta et al. (2020) which can be seen in Figure 1. As we reduce the percentile threshold, we proportionately filter a larger fraction of truly mislabelled data than hard-to-learn samples. Overfitting is another issue that is facilitated by indiscriminately removing samples with low AUM as the proportion of easy-to-learn samples is increased in the filtered dataset (Swayamdipta et al., 2020).

| Sample Id | Text | Label |
|---|---|---|
| 41767 | a damn fine and a truly distinctive and a deeply pertinent film | Negative |
| 42407 | guts and | Negative |
| 62237 | as original and insightful as last week's episode of behind the music . | Negative |
| 6886 | loves the members of the upper class almost as much as they love themselves | Negative |
| 19153 | of how horrible we are to ourselves and each other | Positive |
| 19159 | he script is n't up to the level of the direction | Positive |
| 62178 | though it runs 163 minutes , safe conduct is anything but languorous . | Positive |

Table 3: Filtered examples from Stanford Sentiment Treebank (SST2)

| Sample Id | Text | Label | AUM |
|---|---|---|---|
| 1432 | I disliked the boy's playing the piano loudly. | 0 | -0.501698 |
| 1433 | The boy whose loud playing of the piano I disliked was a student. | 1 | 0.163168 |
| 1434 | The piano which I disliked the boy's playing loudly was badly out of tune. | 0 | -0.349795 |
| 1435 | The boy's loud playing of the piano drove everyone crazy. | 1 | 0.99208 |
| 1436 | The boy's playing the piano loudly drove everyone crazy. | 1 | 0.521766 |
| 1437 | That piano, the boy's loud playing of which drove everyone crazy, was badly out of tune. | 1 | 0.49058 |
| 1438 | That piano, the boy's playing which loudly drove everyone crazy, was badly out of tune. | 0 | -0.333451 |
| 1439 | That piano, which the boy's playing loudly drove everyone crazy, was badly out of tune. | 0 | -0.628104 |
| 1440 | Did that he played the piano surprise you? | 0 | -0.290151 |
| 1441 | Would for him to have played the piano have surprised you? | 0 | -0.430419 |
| 1442 | Is whether he played the piano known? | 0 | -0.292625 |
| 1443 | Did his having played the piano surprise you? | 1 | 0.332026 |

Table 4: Cluster of data points in CoLA with high inter-class similarity; Dominant class - class with samples that have high structural and vocab similarity (This similarity is not quantified numerically but was observed during manual inspection of the data); Class 0 is the major class in this example but the high structural and vocab similarity within class 1 reinforces the modeling process to present it as the dominant class

## 5 Results & Analysis

Table 1 shows the results for Sieving on the real-world dataset (SST-2). This experiment also shows how increasing the percentile threshold decreases the increase in performance, hinting at the fact that large amounts of relevant language information might be getting filtered. Table 5 shows the result for sieving and flipping on synthetic mislabelled samples (SST-2). We expected to observe a drastic dip in performance with noise injection and a relatively large gain once filtered, but we only observed a marginal dip after injecting noise and a marginal increase after filtering in performance. For the Flipping experiment, we only saw ~1% increase after flipping 68 samples (<0.1 percentile threshold) with the lowest AUM. We considered such a low threshold in an attempt to flip only the truly mislabelled data. Investigating further, we saw that about 60-65% of the noise samples were filtered from our experiments. Figure 3 shows the distribution of AUM values of the synthetic noise and clean samples. The graphs clearly show that AUM does help in identifying the mislabelled samples to some extent (Table 2 and Table 3 show the mis-labelled samples we detected in SST-2 and CoLA with low AUM values) but a lot of correctly labeled samples also get filtered depending on how noisy the dataset is. Although there is a high correlation between noise and correctly labeled samples being filtered, the amount of noise alone does not explain this behavior. This leads us to question the efficacy of the AUM metric in NLP datasets.

| Noise % | Accuracy on Unfiltered Data | Accuracy on Filtered Data(Sieving) | Accuracy on Filtered Data(Flipping) |
| --- | --- | --- | --- |
| 20 | 85.38 | **85.94** | 85.83 |
| 40 | 82.70 | 84.26 | **85.96** |

Table 5: Accuracy on Synthetic mislabelled Samples (SST-2); Seed: 100; Threshold Percentile: 90

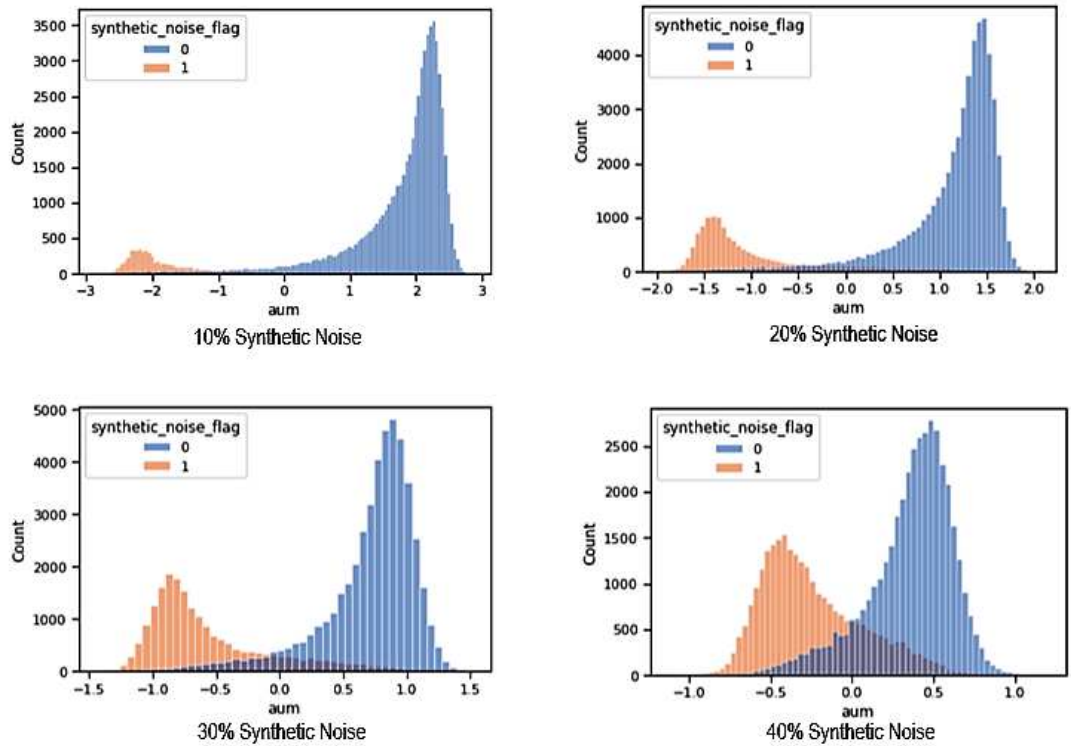

Figure 3: Histogram of AUM values of synthetic noise and unaltered data. (Blue → Unaltered data, Orange → Synthetic noise)

On manual inspection of the CoLA dataset, we found multiple clusters with high feature similarities. Table 4 shows an example of such clusters. We observed that the model is relying on superficial features like word co-occurrence statistics (Sinha et al., 2021), within these clusters and builds a bias for the dominant class label in a particular cluster. Thus the non-dominant class samples (which usually are correctly labeled) get low AUM values instead of the synthetic noise samples. This does not go hand in hand with our previous observations where Figure 3 shows that synthetic noise samples have low AUM, but it is important to note that syn- thetic noise samples also happen to be a part of the non-dominant class in most cases (noise in an acceptable dataset is non-dominant). Again, we emphasize correlation does not imply causation.

In Table 4, the synthetically introduced noise (marked in red) and members of class 0 (marked in yellow) are both parts of the non-dominant class which gives these samples a negative AUM. While the red labeled samples are legitimate candidates for removal, the removal of yellow samples causes loss of correctly labeled data points. We observed the same pattern through all clusters.

# 6 Conclusion

We report on the applicability of AUM on NLP datasets. AUM does help in identifying mislabelled samples available to some extent, but sieving these samples indiscriminately removes large amounts of relevant language information. We hypothesize that the reason AUM works well in image datasets is because of the intrinsic nature of the data samples, i.e., data samples in image datasets have high intra-class and less inter-class feature similarity whereas in NLP datasets, data samples have high inter-class feature similarity as seen in Table 4 and this coupled with the model dependency on superficial features results in low AUM values for the non-dominant class samples instead of the mislabelled class samples consequently reducing the efficacy of the AUM metric in NLP datasets.

## 7 Acknowledgement

We would like to thank Prof. Samuel R. Bowman for his guidance and support. We greatly benefited from the discussions we had with him.

## References

Eyal Beigman and Beata Beigman Klebanov. 2009. Learning with annotation noise. In *Proceedings of the Joint Conference of the 47th Annual Meeting of the ACL and the 4th International Joint Conference on Natural Language Processing of the AFNLP*, pages 280–287, Suntec, Singapore. Association for Computational Linguistics.

Beata Beigman Klebanov, Eyal Beigman, and Daniel Diermeier. 2008. Analyzing disagreements. In *Coling 2008: Proceedings of the workshop on Human Judgements in Computational Linguistics*, pages 2–7, Manchester, UK. Coling 2008 Organizing Committee.

Vikas Bhardwaj, Rebecca Passonneau, Ansaf Saleb-Aouissi, and Nancy Ide. 2010. Anveshan: A framework for analysis of multiple annotators' labeling behavior. In *Proceedings of the Fourth Linguistic Annotation Workshop*, pages 47–55, Uppsala, Sweden. Association for Computational Linguistics.

Samuel R. Bowman, Gabor Angeli, Christopher Potts, and Christopher D. Manning. 2015. A large annotated corpus for learning natural language inference. In *Proceedings of the 2015 Conference on Empirical Methods in Natural Language Processing*, pages 632–642, Lisbon, Portugal. Association for Computational Linguistics.

Isaac Caswell, Julia Kreutzer, Lisa Wang, Ahsan Wahab, Daan van Esch, Nasanbayar Ulzii-Orshikh, Allahsera Tapo, Nishant Subramani, Artem Sokolov, Claytone Sikasote, Monang Setyawan, Supheakmungkol Sarin, Sokhar Samb, Benoît Sagot, Clara Rivera, Annette Rios, Isabel Papadimitriou, Salomey Osei, Pedro Javier Ortiz Suárez, Iroro Orife, Kelechi Ogueji, Rubungo Andre Niyongabo, Toan Q. Nguyen, Mathias Müller, André Müller, Shamsuddeen Hassan Muhammad, Nanda Muhammad, Ayanda Mnyakeni, Jamshidbek Mirzakhalov, Tapiwanashe Matangira, Colin Leong, Nze Lawson, Sneha Kudugunta, Yacine Jernite, Mathias Jenny, Orhan Firat, Bonaventure F. P. Dossou, Sakhile Dlamini, Nisansa de Silva, Sakine Çabuk Ballı, Stella Biderman, Alessia Battisti, Ahmed Baruwa, Ankur Bapna, Pallavi Baljekar, Israel Abebe Azime, Ayodele Awokoya, Duygu Ataman, Orevaoghene Ahia, Oghenefego Ahia, Sweta Agrawal, and Mofetoluwa Adeyemi. 2021. Quality at a glance: An audit of web-crawled multilingual datasets.

Jacob Devlin, Ming-Wei Chang, Kenton Lee, and Kristina Toutanova. 2019. BERT: Pre-training of deep bidirectional transformers for language understanding. In *Proceedings of the 2019 Conference of the North American Chapter of the Association for Computational Linguistics: Human Language Technologies, Volume 1 (Long and Short Papers)*, pages 4171–4186, Minneapolis, Minnesota. Association for Computational Linguistics.

Alex Graves, Abdel-rahman Mohamed, and Geoffrey Hinton. 2013. Speech recognition with deep recurrent neural networks. In *2013 IEEE International Conference on Acoustics, Speech and Signal Processing*, pages 6645–6649. IEEE.

Kaiming He, Xiangyu Zhang, Shaoqing Ren, and Jian Sun. 2015. Deep residual learning for image recognition.

Kalun Ho, Janis Keuper, Franz-Josef Pfreundt, and Margret Keuper. 2021. Learning embeddings for image clustering: An empirical study of triplet loss approaches. In *2020 25th International Conference on Pattern Recognition (ICPR)*, pages 87–94. IEEE.

Ishan Jindal, Daniel Pressel, Brian Lester, and Matthew Nokleby. 2019. An effective label noise model for DNN text classification. In *Proceedings of the 2019 Conference of the North American Chapter of the Association for Computational Linguistics: Human Language Technologies, Volume 1 (Long and Short Papers)*, pages 3246–3256, Minneapolis, Minnesota. Association for Computational Linguistics.

Daniel Kang and Tatsunori B. Hashimoto. 2020. Improved natural language generation via loss truncation. In *Proceedings of the 58th Annual Meeting of the Association for Computational Linguistics*, pages 718–731, Online. Association for Computational Linguistics.

Jingling Li, Mozhi Zhang, Keyulu Xu, John P. Dickerson, and Jimmy Ba. 2020. Noisy labels can induce good representations. ArXiv preprint arXiv:2012.12896.

Tal Linzen. 2020. How can we accelerate progress towards human-like linguistic generalization? In *Proceedings of the 58th Annual Meeting of the Association for Computational Linguistics*, pages 5210–5217, Online. Association for Computational Linguistics.

Ilya Loshchilov and Frank Hutter. 2017. Sgdr: Stochastic gradient descent with warm restarts.

Hassan H. Malik and Vikas S. Bhardwaj. 2011. Automatic training data cleaning for text classification. In *2011 IEEE 11th International Conference on Data Mining Workshops*, pages 442–449.

Valentin Malykh. 2019. Robust to noise models in natural language processing tasks. In *Proceedings of the 57th Annual Meeting of the Association for Computational Linguistics: Student Research Workshop*, pages 10–16, Florence, Italy. Association for Computational Linguistics.

Nicolas M. Müller and Karla Markert. 2019. Identifying mislabeled instances in classification datasets. In *2019 International Joint Conference on Neural Networks (IJCNN)*, pages 1–8.

Ellie Pavlick and Tom Kwiatkowski. 2019. Inherent disagreements in human textual inferences. *Transactions of the Association for Computational Linguistics*, 7:677–694.

Geoff Pleiss, Tianyi Zhang, Ethan R. Elenberg, and Kilian Q. Weinberger. 2020. Identifying mislabeled data using the area under the margin ranking.

Pranav Rajpurkar, Jian Zhang, Konstantin Lopyrev, and Percy Liang. 2016. SQuAD: 100,000+ questions for machine comprehension of text. In *Proceedings of the 2016 Conference on Empirical Methods in Natural Language Processing*, pages 2383–2392, Austin, Texas. Association for Computational Linguistics.

Victor Sanh, Lysandre Debut, Julien Chaumond, and Thomas Wolf. 2019. Distilbert, a distilled version of bert: smaller, faster, cheaper and lighter. *arXiv preprint arXiv:1910.01108*.

Koustuv Sinha, Robin Jia, Dieuwke Hupkes, Joelle Pineau, Adina Williams, and Douwe Kiela. 2021. Masked language modeling and the distributional hypothesis: Order word matters pre-training for little. *arXiv preprint arXiv:2104.06644*.

Richard Socher, Alex Perelygin, Jean Y Wu, Jason Chuang, Christopher D Manning, Andrew Y Ng, and Christopher Potts. 2013. Recursive deep models for semantic compositionality over a sentiment treebank. In *Proceedings of the conference on empirical methods in natural language processing (EMNLP)*, volume 1631, page 1642. Citeseer.

L. Venkata Subramaniam, Shourya Roy, Tanveer A. Faruquie, and Sumit Negi. 2009. A survey of types of text noise and techniques to handle noisy text. In *Proceedings of The Third Workshop on Analytics for Noisy Unstructured Text Data*, AND '09, page 115–122, New York, NY, USA. Association for Computing Machinery.

Swabha Swayamdipta, Roy Schwartz, Nicholas Lourie, Yizhong Wang, Hannaneh Hajishirzi, Noah A. Smith, and Yejin Choi. 2020. Dataset cartography: Mapping and diagnosing datasets with training dynamics. In *Proceedings of the 2020 Conference on Empirical Methods in Natural Language Processing (EMNLP)*, pages 9275–9293, Online. Association for Computational Linguistics.

Alex Wang, Amanpreet Singh, Julian Michael, Felix Hill, Omer Levy, and Samuel R. Bowman. 2019. Glue: A multi-task benchmark and analysis platform for natural language understanding.

Alex Warstadt, Amanpreet Singh, and Samuel R. Bowman. 2019. Neural network acceptability judgments. *Transactions of the Association for Computational Linguistics*, 7:625–641.

Hainan Xu and Philipp Koehn. 2017. Zipporah: a fast and scalable data cleaning system for noisy web-crawled parallel corpora. In *Proceedings of the 2017 Conference on Empirical Methods in Natural Language Processing*, pages 2945–2950, Copenhagen, Denmark. Association for Computational Linguistics.

Chiyuan Zhang, Samy Bengio, Moritz Hardt, Benjamin Recht, and Oriol Vinyals. 2016. Understanding deep learning requires rethinking generalization. *CoRR*, abs/1611.03530.

Yivan Zhang and Masashi Sugiyama. 2021. Approximating instance-dependent noise via instance-confidence embedding.